\pdfoutput=1
\documentclass[11pt]{article}
\usepackage{acl}

\usepackage{times}
\usepackage{latexsym}
\usepackage[T1]{fontenc}
\usepackage[utf8]{inputenc}
\usepackage{microtype}
\usepackage{inconsolata}
\usepackage{graphicx}
\usepackage{amsmath}
\usepackage{amsfonts}
\usepackage{booktabs}
\usepackage{multirow}

\usepackage{tcolorbox}

\usepackage{xcolor}         
\usepackage{color, colortbl}

\definecolor{backred}{RGB}{255, 190, 190}
\definecolor{backblue}{RGB}{210, 230, 250}

\definecolor{verylightgray}{gray}{0.95}

\newcommand\bench{InterFeedback-Bench}
\newcommand\agent{InterFeedback}
\newcommand\dataset{InterFeedback-Human}

\title{\agent{}: Unveiling Interactive Intelligence of Large Multimodal Models with Human Feedback}

\author{
  Henry Hengyuan Zhao\thanks{Equal Contribution.},
  Wenqi Pei\footnotemark[1],
  Yifei Tao\footnotemark[1], \\
  \textbf{Haiyang Mei},
  \textbf{Mike Zheng Shou}\thanks{Corresponding author.} \\ \\
  Show Lab, National University of Singapore
}

\begin{document}
\maketitle

\begin{abstract}

Existing benchmarks do not test Large Multimodal Models (LMMs) on their interactive intelligence with human users, which is vital for developing general-purpose AI assistants. We design \agent{}, an interactive framework, which can be applied to any LMM and dataset to assess this ability autonomously. On top of this, we introduce \bench{} that evaluates interactive intelligence using two representative datasets, MMMU-Pro and MathVerse, to test 10 different open-source LMMs. Additionally, we present \dataset{}, a newly collected dataset of 120 cases designed for manually testing interactive performance in proprietary models such as OpenAI-o1 and Claude-Sonnet-4. Our evaluation results indicate that even the state-of-the-art LMM, OpenAI-o1, struggles to refine its responses based on human feedback, achieving an average score of less than 50\%. Our findings point to the need for methods that can enhance LMMs' capabilities to interpret and benefit from feedback.
\end{abstract}

\section{Introduction}

\begin{figure}[t]
    \centering
    \includegraphics[width=0.76\linewidth]{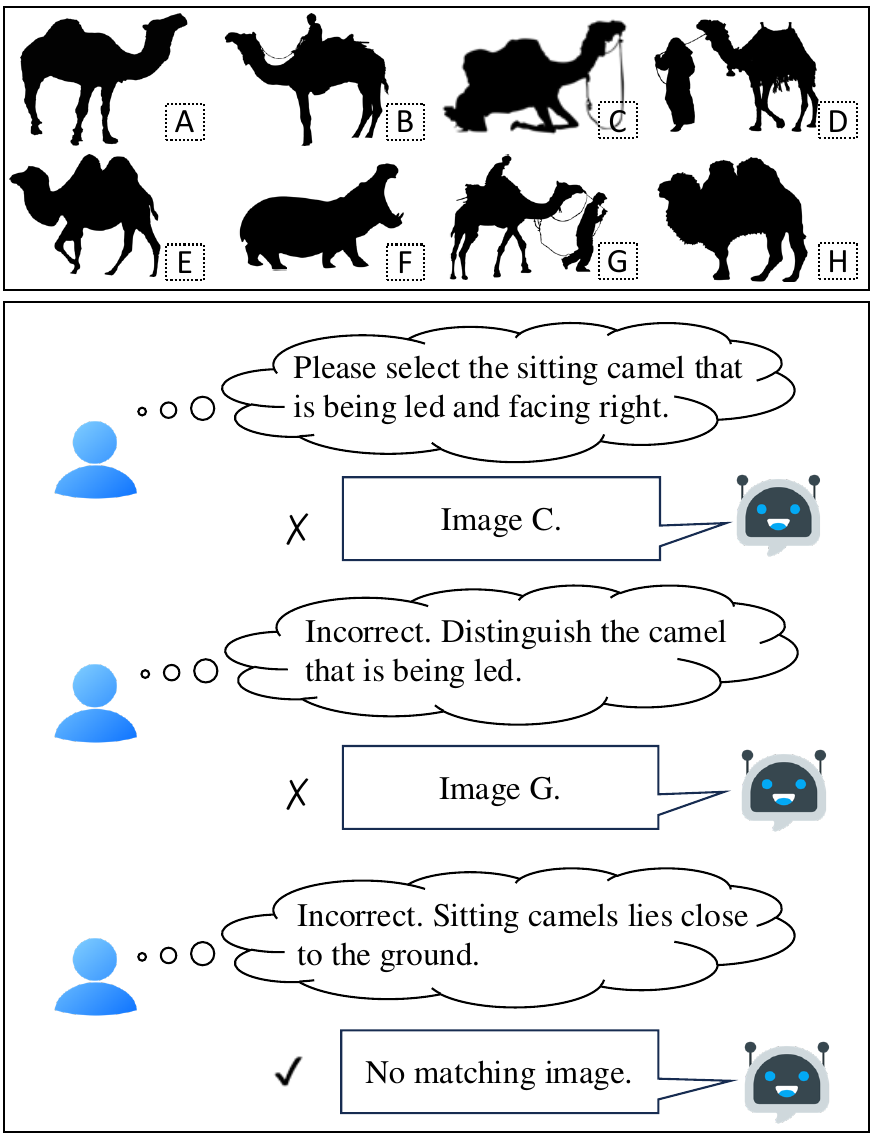}
    \caption{Illustration of an interactive feedback scenario. When models generate incorrect responses, human users provide pertinent feedback to interactively refine the answers.}
    \label{fig:teaser}
\end{figure}

In this paper, we are curious about the question \textit{\textbf{``Can Large Multimodal Models evolve through Interactive Human Feedback?''}} It is central to developing general-purpose AI assistants with Large Multimodal Models (LMMs). While these models show exceptional performance on tackling multimodal tasks directly, their ability to interact with humans remains largely unknown. We argue that an LMM functioning as the general assistant should possess two capabilities: 1) exceptional problem-solving ability and 2) the ability to improve itself through feedback, as shown in Figure \ref{fig:teaser} (e.g., human feedback, execution results). In this work, we focus on the latter capability, which has been rarely examined in existing benchmarks.

Humans are highly adaptive, continuously refining their skills through feedback--a fundamental process for acquiring knowledge and solving problems. Likewise, advanced LMM models should be designed to learn from feedback, ensuring better alignment with real-world needs and enhancing their problem-solving capabilities in Human-AI Interaction (HAI). Recently, a surge of large multimodal models (LMMs) \citep{gpt4o, qwen2-vl, molmo, zhao2024genixer, llava-onevision, zhao2024lova, chen2024internvl} have been developed to handle various tasks, including general vision-language understanding~\citep{liu2023mmbench, li2023seedbench}, expert-level multimodal understanding~\citep{yue2024mmmu, mmmupro}, and scientific reasoning~\citep{lu2022learn_scienceqa, lu2024mathvista, mathverse}. However, these LMMs are primarily tested in a static way, overlooking their great potential in an interactive process such as interactive coding~\citep{jimenez2024swebench, swebenchmm}, computer usage~\citep{worldgui, lin2024videogui, assistgui, OSWorld}, and clinical reasoning~\citep{li2024mediq}. Consequently, the interactive intelligence of LMMs remains largely unexplored, and the development of a standard benchmark for evaluating their interactive intelligence remains an open challenge.

The key challenge in evaluating the interactive intelligence of LMMs is the automatic model tests. In practice, for the same query, different LMMs often produce varied responses, necessitating that humans offer tailored feedback for each conversation round.
To address this issue, we propose \textbf{\agent{}} a straightforward problem-solving framework that enables any LMM to tackle multimodal tasks interactively by leveraging proprietary models such as GPT-4o~\citep{gpt4o} to simulate humans, inspired in previous studies~\citep{yao2025taubench, chatshop, yoon-etal-2024-evaluating, VideoAutoArena}.

On top of this framework, we present \textbf{\bench{}}, a benchmark designed to comprehensively evaluate LMMs for two purposes: 1) \textbf{the ability to interactively solve problems} and 2) \textbf{the capability of interpreting the feedback to improve themselves}. We demonstrate with two challenging pre-existing datasets: MMMU-Pro~\citep{mmmupro} and Mathverse~\citep{mathverse}. Additionally, for a more in-depth investigation, we conduct human evaluation on four closed-source proprietary models: GPT-4o \citep{gpt4o}, OpenAI-o1 \citep{openaio1}, Claude-3.5-Sonnet \citep{claude3.5}, and Gemini-2.0 \citep{gemini-2.0} with a trained user acting as the feedback provider. Finally, we manually collected a dataset \textbf{\dataset{}} containing 120 samples for this assessment.

Our experimental results reveal several compelling insights: 1) Interactive process could improve the performance of most LMMs in solving challenging problems; 2) Existing LMMs exhibit suboptimal performance in interpreting and incorporating feedback; 3) Accuracy result may not truly reflect the model’s capability to improve itself from feedback; 4) High-quality feedback is essential, as subpar feedback can degrade performance even more than a simple binary (0/1) correctness signal; 5) LMM may not truly reasoning, we find out that LMMs resort to guessing answer even on a simple question according to human. These findings point to the need for methods that can enhance the LMM's capability to interpret and benefit from feedback. In summary, our contributions are:
\begin{itemize}
    \item We take the first step toward exploring the interactive intelligence of LMMs in improving themselves through human feedback.
    \item We propose a straightforward and extensible framework \agent{} which allows any LMM to interactively solve problems.
    \item We construct \bench{}, a novel and universal benchmark for assessing the ability of interactive problem-solving of LMMs. 
    \item We conduct comprehensive evaluations and in-depth analysis, providing several compelling insights for future model alignment.
\end{itemize}

\section{Related Work}
\textbf{Large Multimodal Models.}
The LLaVA-series works~\citep{liu2023llava, llava1.5, liu2024llavanext, llava-onevision} demonstrate that training with supervised fine-tuning (SFT) multimodal data and expand the vision lens would produce compatible multimodal reasoning ability. By adopting a large-scale image-text corpus for instruction tuning, Qwen2-VL~\citep{qwen2-vl}, CogVLM~\citep{wang2023cogvlm}, InternVL2 \citep{internvl2} have achieved exceptional performance on various multimodal abilities. Moreover, Molmo \citep{molmo} proposes to train an LMM from scratch with only the human-annotated data. Unlike these large models, MiniCPM-V \citep{yao2024minicpmvgpt4vlevelmllm} and Phi-3.5-Vision \citep{phi3model} propose to train lightweight yet SOTA LMMs. Despite their exceptional performance on multimodal benchmarks of varying difficulty, such as MMMU-Pro \citep{mmmupro} and MathVista \citep{lu2024mathvista}, it remains unclear how well these LMMs demonstrate interactive intelligence in Human-AI Interaction scenarios. In this paper, we conduct the evaluation of these LMMs to explore this basic yet vital capability (i.e., evolve through interactive human feedback).

\begin{figure*}[!t]
    \centering
    \includegraphics[width=0.78\linewidth]{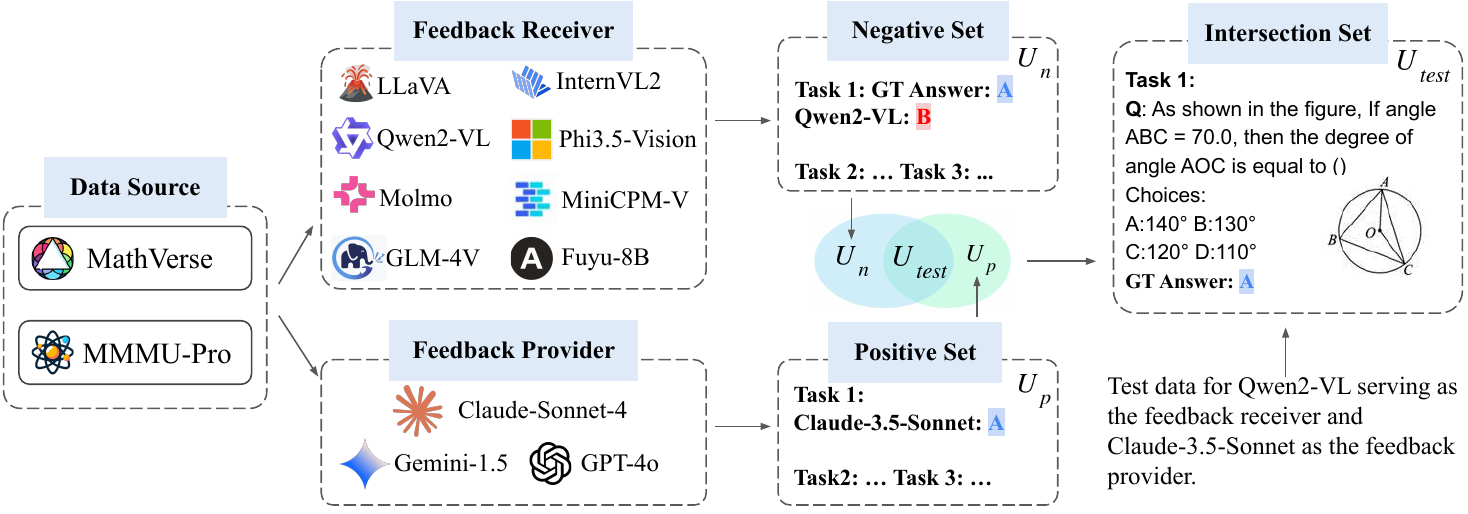}
    \vspace{-1mm}
    \caption{Overview of the test data construction process for \bench{}. For each LMM serving as the feedback receiver, we process each instance from a target dataset (e.g., MathVerse) and collect the error cases to form a negative set. The feedback provider then processes the same instances to build a positive set. Finally, we curate test data by selecting the intersection of both sets.}
    \label{fig:dataconstruct}
\end{figure*}

\textbf{Multimodal Benchmarks.}
Traditional vision-language benchmarks focus on visual question answering~\citep{VQAv2}, image captioning~\citep{cococaption}, as well as other benchmarks for specialized scenarios such as scene text understanding~\citep{textvqa}, commonsense reasoning~\citep{zellers2019recognition}, outside knowledge~\citep{marino2019ok_okvqa,AOKVQA}. The recent development of LMM posts a strong need for modernized multimodal benchmarks~\citep{worldgui,liu2023mmbench,li2023seedbench,yu2023mmvet, yue2024mmmu, lu2024mathvista, mathverse, li2024fire} such as MMBench~\citep{liu2023mmbench}, MMMU-pro \citep{mmmupro}, and MathVerse \citep{mathverse} which involve comprehensively evaluating current LMMs on various multimodal abilities. However, these benchmarks primarily focus on static testing processes, overlooking the interactive testing process that is vital in human-AI interaction scenarios.

\textbf{Human-AI Interaction.}
Investigating how humans and AI systems communicate and collaborate is critical for shaping applications such as virtual assistants \citep{virvou2022emerging}, personalized recommendations \citep{dodeja2024towards}, autonomous vehicles \citep{zhang2021human}, and healthcare diagnostics \citep{mckinney2020international}. Recent LLMs-driven techniques, such as memory~\citep{park2023generative} and iterative~\citep{zhang2023human} mechanisms offer expert-level collaboration. While LMMs \citep{molmo,qwen2-vl} excel in multimodal tasks, their potential for HAI problem-solving \citep{swebenchmm, li2024mediq} remains underexplored. By offering a unified framework and meticulously curated data, our \bench{} enables evaluation of LMMs on these capabilities and lays a foundation for advancing multimodal HAI problem-solving.

\textbf{User Stimulation with LLM.} Recently, previous work in order to build multi-agent system \citep{khan2024debatingpersuasivellmsleads}, stimulate human-AI interaction \citep{yao2025taubench}, evaluate LMMs in video analysis \citep{VideoAutoArena}, stimulate real users in a web shopping scenario \cite{chatshop}, evaluate the conversational recommender systems \citep{yoon-etal-2024-evaluating} determine whether to use LLM or LMM to stimulate the user. However, previous works have overlooked the importance of ensuring the reliability of LLMs or LMMs that are used to stimulate the users. In this paper, we curate test data by selecting only the samples that LMMs correctly address, minimizing unreliable interaction results.

\section{\bench{}}
In this section, we first introduce the automated interactive benchmarking process in Section \ref{sec:interbench}. We begin by formulating the concept of interactive problem-solving, followed by a discussion of the data curation process. We then present the proposed interactive framework, \agent{}. Next, in Section \ref{sec:humanbench}, we elaborate on the human benchmarking component, detailing the data collection and the proposed feedback providing strategy.

\subsection{Automated Interactive Benchmarking}
\label{sec:interbench}
\subsubsection{Formulation}
The \bench{} formalizes the interactive problem-solving process with feedback in a partially observable Markov decision process (POMDP) $(\mathcal{S}, \mathcal{O}, \mathcal{A}, \mathcal{T}, \mathcal{R})$ with state space $\mathcal{S}$, observation $\mathcal{O}$, action space $\mathcal{A}$, transition function $\mathcal{T}$: $\mathcal{S} \times \mathcal{A} \rightarrow \mathcal{S}$, and reward function $\mathcal{R}$: $\mathcal{S} \times \mathcal{A} \rightarrow \mathbb{R}$. In our setting, given a natural language question $q$ (e.g., \textit{Please select the sitting camel that
is being led and facing right}) and the input image $v$, the model first gets the observation $o_{t} \in \mathcal{O}$ from the state $s_t \in \mathcal{S}$ in the execution environment and then generate the action $a_t \in \mathcal{A}$. The $a_t$ is the response from models in natural language. The reward function $\mathcal{R}$: $\mathcal{S} \times \mathcal{A} \rightarrow \{0, 1\}$ here returns a binary value indicating the task correctness status. It is implemented by the exact match: returning 1 if the predicted answer exactly matches the ground-truth, and 0 otherwise. The observation $o_t$ includes both the correctness signal from the reward function and the feedback from the humans.

\subsubsection{Data Curation}
\textbf{Data sources.}
To ensure the quality and difficulty of multimodal tasks, inspired by previous benchmarks demonstrated on pre-existing datasets \citep{yang2023intercode, li2024vlrewardbenchchallengingbenchmarkvisionlanguage}, we choose to test LMMs on two challenging datasets: MathVerse \citep{mathverse} and MMMU-Pro \citep{mmmupro}. MathVerse is a visual math benchmark that includes various mathematical problems, and 3,940 samples (testmini set) are used in our work. MMMU-Pro is a comprehensive multimodal benchmark and we use 1,730 expert-level questions (single image mode). Both datasets are challenging even for the model GPT-4o which achieves only 64.7\% accuracy on MMMU-Pro (Standard 4 options). 

\begin{figure*}[t]
    \centering
    \includegraphics[width=0.8\linewidth]{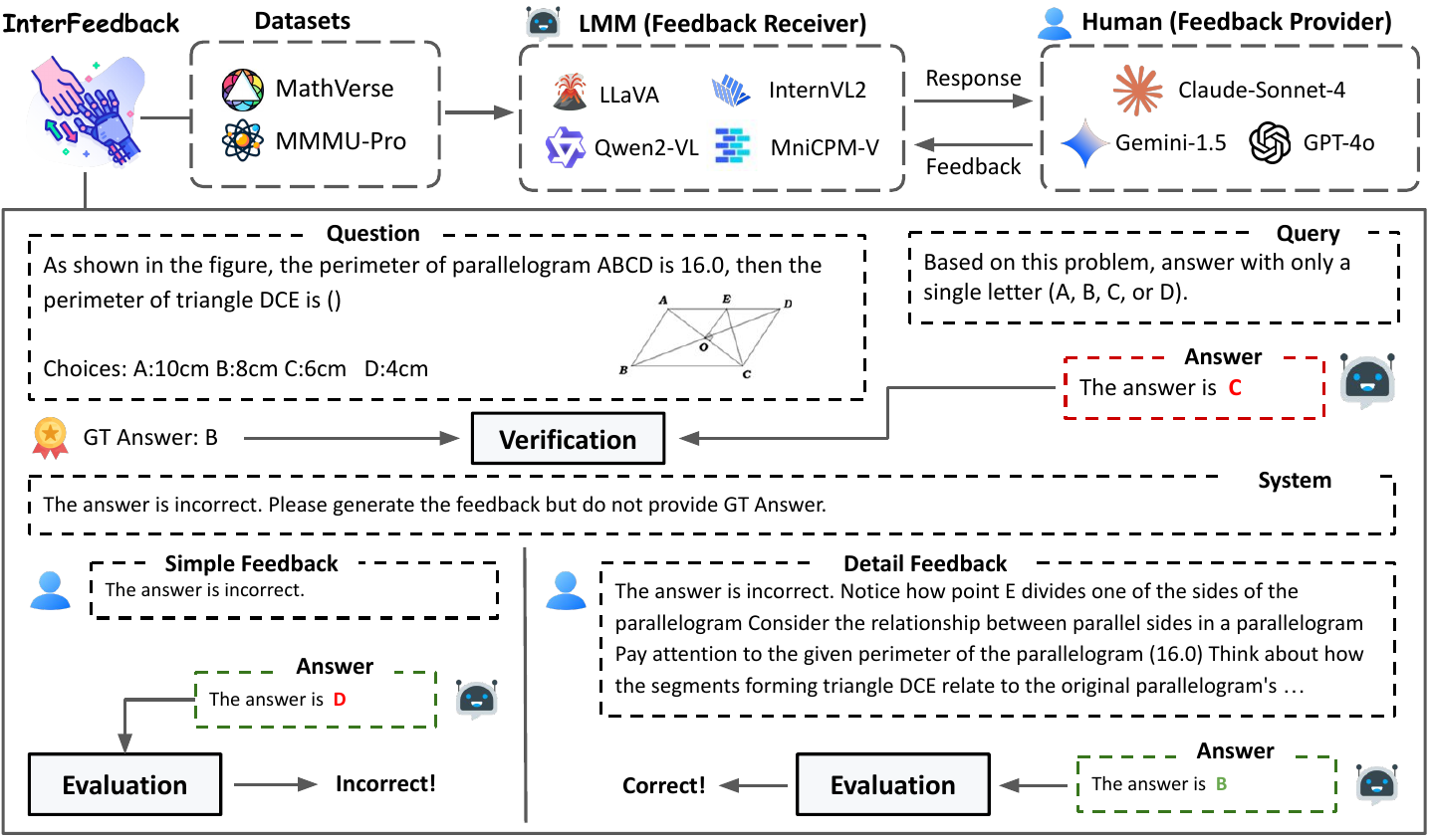}
    \caption{Overview of the proposed framework \agent{} for assessing an LMM's ability to improve itself through feedback. The model interacts with humans to progressively solve a problem, and after each conversation round, we verify the correctness of the answer. If the answer is incorrect, an LMM-stimulated human will provide constructive feedback. We implement two types of feedback to investigate the behavior of LMMs.}
    \label{fig:interfeedbackprocess}
\end{figure*}

\textbf{Data selection process.}
We choose to use proprietary LMMs, such as GPT-4o, for stimulating the humans to give feedback mimicking human-AI interactions. The primary challenge, however, is ensuring that the feedback generated by these models is reliable, as even models like GPT-4o and Claude-Sonnet-4 still do not perform correctly on all test samples. Therefore, we construct the test data by selecting the intersection set that feedback provider $M_p$ solves correctly while $M_r$ does not, as shown in Figure \ref{fig:dataconstruct}. Specifically, the pipeline includes three parts: 1) feedback receiver LMM locally running; 2) feedback provider LMM API-calling; and 3) intersection set selection. Such a data construction process leads to each tested LMM having a different test data set.

Specifically, given a test dataset \(D\), we begin by having the feedback receiver model \(M_r\) process every instance in \(D\) to produce a negative set \(U_n\) consisting of tasks it fails to solve correctly. Next, the feedback provider model \(M_p\) processes the same dataset to generate a positive set \(U_p\) comprising tasks it solves correctly. We then define \(U_{\text{test}}\) as the intersection of \(U_n\) and \(U_p\), i.e.,
\[
U_{\text{test}} = U_n \cap U_p,
\]
which means that \(U_{\text{test}}\) contains tasks that \(M_p\) solves correctly but \(M_r\) does not. This approach ensures that the feedback generated by \(M_p\) is both relevant and reliable.

\subsubsection{\agent{} Framework} 
To enable an interactive problem-solving process, we propose a new straightforward framework \textbf{\agent{}}. It includes two roles: feedback receiver $M_r$ and feedback provider $M_p$, as shown in Figure \ref{fig:interfeedbackprocess}. The feedback receiver is the candidate LMMs (e.g., Qwen2-VL) ready for the benchmark and the feedback provider is the SOTA LMM (e.g., GPT-4o) for providing the pertinent feedback in each time step in place of a human. Consider at timestep $t$, the output of $M_r$ is $a_t$, and the feedback provider $M_p$ has to follow the policy that provides the feedback $f_t$ from the mapping $: F(a_t, s_t) \rightarrow f_t$. The $s_t$ denotes the correctness signal from the verification process via the reward function. We record the model outputs for the final evaluation.

\textbf{Feedback types.} Additionally, we propose a simplified feedback mechanism that only indicates correctness (i.e., correct or incorrect), without a detailed explanation. In summary, we evaluate the models using two feedback types: \emph{Detail} and \emph{Simple}. The \emph{Detail} feedback comprises both \emph{Simple} feedback and detailed LMM-generated explanation.

\subsection{Human Benchmarking}
\label{sec:humanbench}
In the previous section, we employed proprietary LMMs as feedback providers. Naturally, how well do these models perform when receiving feedback? We begin to assess the proprietary LMMs with a human-in-the-loop process. The feedback provider $M_p$ is a trained user who fully understands all the questions in the newly curated dataset \dataset{}. The feedback receiver $M_r$ is the proprietary LMMs including OpenAI-o1, GPT-4o, Gemini-2.0, and Claude-3.5-Sonnet. This evaluation aims to assess how effectively these proprietary models can serve as assistants in a human-AI interaction system.

\subsubsection{Data Collection.} We gather challenging data examples across diverse domains: visual logic, mathematics, and coding. These were selected to probe the cognitive depth of the models, especially when confronted with complex, multi-step reasoning problems. The visual logic data is manually collected from publicly available resources. The emphasis on visual logic tasks reflects the growing demand for models to handle image-based reasoning challenges, such as pattern recognition \citep{wei2025slowperceptionletsperceive} (e.g., determining the next shape in a sequence) and character-based logic (e.g., interpreting transformations between symbols). We also collect the multimodal mathematics data from the existing dataset MathVerse \citep{mathverse} and the multimodal expert-level data from MMMU-Pro \citep{mmmupro}. Additionally, we also involve the natural language task into \dataset{} to analyze such capability in the NLP area.

In summary, \dataset{} encompasses a total of 120 tasks distributed across the five task types: 80 visual logic tasks, 10 mathematical logic tasks (sampled from NuminaMath \citep{numina_math_datasets}), 10 coding tasks (sampled from CodeComprehension \citep{imbuecodedataset}), 10 MMMU-Pro tasks, and 10 MathVerse tasks.

\subsubsection{Hierarchical Feedback} We design a hierarchical feedback generation scheme to gradually increase the information intensity. Specifically, we ask the human to give the following three-level feedback:

\begin{itemize}
\item Level 1: Provide a basic and simple description that leads to the correct answer.
\item Level 2: Provide an expanded explanation that leads to the correct answer.
\item Level 3: The correct answer is \underline{GT Answer}. Provide a comprehensive and detailed explanation that leads to the correct answer.
\end{itemize}

Since most of our questions have four options, giving more than three rounds of feedback might let the model guess the answer by elimination rather than by reasoning. For example, if the correct answer is A and the model has already given B, C, and D, a third round of feedback is unnecessary. Therefore, we directly provide the \underline{GT Answer} in Level 3 feedback prompts to test the models' ability to explain their thinking process.

\subsubsection{Evaluation Integration} To ensure fairness and consistency in our evaluation, we engaged only one experienced user. Since human-in-the-loop feedback is inherently subjective, involving multiple participants could introduce variability due to differences in background and expertise. This approach helps maintain the reliability of the relative performance comparisons across candidate LMMs.

\begin{table*}[!t]
\centering
\tabcolsep=1.5mm
\renewcommand{\arraystretch}{1.1}
\scalebox{0.64}{
\begin{tabular}{lcc|>{\columncolor{backblue!50}}c|cc|>{\columncolor{backblue!50}}c|cc|>{\columncolor{backblue!50}}c|cc}
\toprule
\multirow{2}{*}{Model} & & & \multicolumn{3}{c|}{GPT-4o} & \multicolumn{3}{c|}{Gemini-1.5-Flash} & \multicolumn{3}{c}{Claude-Sonnet-4}\\
\cmidrule{4-12}
& Acc (\%) & \# Neg & \# Test & Detail (\%) & Simple (\%) & \# Test & Detail (\%) & Simple (\%) & \# Test & Detail (\%) & Simple (\%)\\
\midrule
\multicolumn{12}{c}{Non-Thinking Models}\\
\hline
LLaVa-OneVision-7B & 25.6 & 2933 & 373 & 36.2 & 18.0 & 428 & 29.0 & 15.7 & 820 & 38.3 & 23.8 \\
Molmo-7B & 25.6 & 2931 & 452 & 55.1 & 52.0 & 507 & 36.5 & 38.9 & 987 & 15.3 & 34.3 \\
MiniCPM-V & 16.2 & 3301 & 552 & 28.4 & 20.3 & 741 & 16.6 & 25.4 & 1195 & 5.3 & 12.1\\
GLM-4V-9B & 20.2 & 3146 & 440 & 38.6 & 28.2 & 568 & 30.1 & 29.9 & 1015 & 22.9 & 21.9 \\
Phi3.5-Vision-4.2B & 19.0 & 3192 & 534 & 36.1 & 33.7 & 579 & 31.3 & 33.7 & 1045 & 21.1 & 24.2 \\
LLaVa-1.5-7B & 13.5 & 3409 & 763 & 23.2 & 14.3 & 678 & 18.0 & 14.7 & 1256 & 3.3 & 5.8 \\
LLaVa-1.6-Mistral-7B & 14.8 & 3357 & 549 & 41.0 & 35.9 & 661 & 5.9 & 5.9 & 1212 & 14.8 & 17.7\\
Fuyu-8B & 21.8 & 3083 & 582 & 24.1 & 19.8 & 635 & 15.0 & 12.9 & 1187 & 17.9 & 15.5\\
InternVL2-8B & 38.1 & 2440 & 379 & 49.6 & 41.2 & 375 & 48.8 & 44.4 & 547 & 21.4 & 26.7\\
Qwen2-VL-7B & 22.5 & 3052 & 295 & 66.8 & 72.2 & 470 & 41.9 & 44.9 & 774 & 34.4 & 35.8 \\
Qwen2.5-VL-7B & 31.5 & 2698 & 266 & 69.2 & 62.4 & 350 & 45.4 & 42.6 & 1521 & 46.8 & 43.9 \\
\hline
\multicolumn{12}{c}{Thinking Models}\\
\hline
Seed-1.5-VL-Thinking & 47.4 & 2072 & 73 & 67.1 & 63.0 & 70 & 64.3 & 58.6 & 474 & 88.6 & 90.5\\
\bottomrule
\end{tabular}
}
\caption{\textbf{Correction Rate Results of three Feedback Providers on MathVerse Dataset.} \textbf{Acc (\%)}: The average accuracy of MathVerse's \emph{testmini} set. (Calculated by our prompt template.) The results are tested by ourselves. \textbf{\# Neg}: The number of negative samples produced by the model. \textbf{\# Test}: The total number of test samples evaluated. \textbf{Detail (\%)}: correction rate of using LMM-generated feedback. \textbf{Simple (\%)}: correction rate of using simple feedback (0 or 1).}
\label{tab:mathverse}
\end{table*}

\begin{table*}[t]
\centering
\tabcolsep=1.5mm
\renewcommand{\arraystretch}{1}
\scalebox{0.66}{
\begin{tabular}{lcc|>{\columncolor{backblue!50}}c|cc|>{\columncolor{backblue!50}}c|cc|>{\columncolor{backblue!50}}c|cc}
\toprule
\multirow{2}{*}{Model} & & & \multicolumn{3}{c|}{GPT-4o} & \multicolumn{3}{c|}{Gemini-1.5-Flash} & \multicolumn{3}{c}{Claude-Sonnet-4}\\
\cmidrule{4-12}
& Acc (\%) & \# Neg & \# Test & Detail (\%) & Simple (\%) & \# Test & Detail (\%) & Simple (\%) & \# Test & Detail (\%) & Simple (\%)\\
\midrule
\multicolumn{12}{c}{Non-Thinking Models}\\
\hline
LLaVa-OneVision-7B & 47.1 & 915 & 312 & 31.7 & 15.7 & 333 & 35.4 & 18.6 & 539 & 42.2 & 30.6\\
Molmo-7B & 43.8 & 973 & 362 & 51.7 & 48.9 & 383 & 41.5 & 43.1 & 593 & 19.7 & 33.9\\
MiniCPM-V & 38.1 & 1071 & 410 & 27.3 & 23.7 & 503 & 21.5 & 21.7 & 688 & 7.0 & 15.3\\
GLM-4V-9B & 46.0 & 935 & 327 & 38.8 & 30.0 & 359 & 38.7 & 31.5 & 577 & 27.6 & 23.6\\
Phi3.5-Vision-4.2B & 43.2 & 983 & 366 & 44.3 & 42.3 & 396 & 40.9 & 39.6 & 611 & 31.8 & 31.1 \\
LLaVa-1.5-7B & 36.5 & 1099 & 506 & 31.9 & 12.3 & 470 & 20.0 & 16.0 & 720 & 8.6 & 11.8\\
LLaVa-1.6-Mistral-7B & 38.8 & 1058 & 432 & 46.1 & 36.1 & 429 & 14.7 & 14.7 & 682 & 27.3 & 25.5 \\
Fuyu-8B & 34.1 & 1140 & 481 & 6.0 & 8.7 & 1140 & 3.7 & 3.5 & 768 & 10.2 & 8.7\\
InternVL2-8B & 45.7 & 939 & 343 & 50.1 & 41.4 & 329 & 57.1 & 50.2 & 435 & 23.7 & 32.9\\
Qwen2-VL-7B & 48.1 & 898 & 268 & 50.4 & 44.8 & 322 & 39.4 & 37.6 & 525 & 35.6 & 33.7\\
Qwen2.5-VL-7B & 50.0 & 865 & 839 & 39.6 & 36.1 & 323 & 44.9 & 39.0 & 839 & 39.6 & 36.1\\
\midrule
\multicolumn{12}{c}{Thinking Models}\\
\hline
Seed-1.5-VL-Thinking & 94.2 & 101 & 20 & 80.0 & 70.0 & 31 & 64.5 & 64.5 & 50 & 74.0 & 70.0\\
\bottomrule
\end{tabular}
}
\caption{\textbf{Correction Rate Results of three Feedback Providers on MMMU-Pro Dataset.} We test models on a single image setting of MMMU-Pro.}
\label{tab:mmmupro}
\end{table*}

\section{Experiments}

\subsection{Experiment Setup}

\textbf{Evaluation Models.} We evaluate the performance of foundation models served as the feedback receiver $M_r$ across 12 representative LMMs: LLaVA-1.5-7B \cite{llava1.5}, LLaVA-1.6-7B \cite{liu2024llavanext} (Mistral-7B), LLaVa-OneVision-7B \cite{llava-onevision} (Qwen2-7B \cite{qwen2}), Qwen2-VL-7B \cite{qwen2-vl}, Qwen2.5-VL-7B \cite{qwen2.5-VL}, GLM-4V-9B \cite{wang2023cogvlm}, InternVL2 \cite{internvl2}, Molmo \cite{molmo}, MiniCPM-V \cite{yao2024minicpmvgpt4vlevelmllm}, Phi-3.5-Vision \cite{phi3model}, Fuyu-8B \cite{fuyu-8b}, and Seed-1.5-VL-Thinking \footnote{doubao-1-5-thinking-vision-pro-250428} \cite{seed1_5vl}. The feedback provider $M_p$ includes the three best available models from three model families: OpenAI (\texttt{gpt-4o-2024-08-06}), Gemini (\texttt{Gemini-1.5-Pro}), and Claude (\texttt{claude-sonnet-4-20250514}).

\textbf{Evaluation Metrics.} In addition to the \textbf{Accuracy} metric, we leverage the \textbf{Correction Rate}, defined as the percentage of corrected answers of all erroneous samples. Let $N$ denote the total number of samples, $N_e$ the number of erroneous samples, and $N_c$ the number of samples that have been corrected. The Accuracy and Correction Rate metrics can be formulated as follows:
\begin{align}
\text{Accuracy} &=  \frac{N-N_e}{N} \times 100\%, \\
\text{Correction}~&\text{Rate} =  \frac{N_c}{N_e} \times 100\%.
\end{align}

\textbf{Implementation Details.} We set the temperature to 0 for all tested models and API models. The image resolution of the Qwen2-VL model we restrict to 512 $\times$ 512 to avoid the memory exceeded error. All evaluations were conducted on two NVIDIA RTX A6000 GPUs. To ensure the reliability of results, we obtain the intersection set for both the feedback receiver and provider models that are able to output the correct answer format. Based on our preliminary experiments, we limited the interactive benchmarking to a single round. This decision is driven by two observations: most models fail to provide correct answers in subsequent rounds, and multiple rounds tend to lead to answer guessing, which undermines the reliability of quantitative evaluation.

\textbf{Feedback Types.} As introduced in Section~\ref{sec:interbench}, we employ proprietary LMMs to stimulate the human to provide pertinent feedback at each conversation round. Additionally, we propose a simplified feedback mechanism that only indicates correctness (i.e., correct or incorrect), without a detailed explanation. In summary, we evaluate the models using two feedback types: \emph{Detail} and \emph{Simple}. The \emph{Detail} feedback comprises both \emph{Simple} feedback and detailed LMM-generated feedback.

\subsection{Experimental Analysis on Interactive Benchmarking}

To thoroughly investigate the ability of LMMs to integrate feedback and improve their problem-solving performance, we present evaluation results for various models on two datasets—MathVerse \citep{mathverse} in Table~\ref{tab:mathverse} and MMMU-Pro \citep{mmmupro} in Table~\ref{tab:mmmupro}, respectively. Below, we provide a detailed discussion of key findings.

\textbf{Whether interactive process improves the performance of LMMs? Yes.} As demonstrated in both tables, integrating our proposed framework \agent{} enables most models to benefit from feedback provided by SOTA LMMs, such as GPT-4o and Claude-Sonnet-4. Notably, even the weaker model Fuyu-8B sees 24.1\% of its erroneous samples corrected through GPT-4o's feedback.

\begin{table*}[!t]
\centering
\tabcolsep=1.2mm
\renewcommand{\arraystretch}{1.1}
\scalebox{0.96}{
\begin{tabular}{l|ccccc|c}
\toprule
\multirow{1}{*}{Model} & \multicolumn{1}{c}{Visual Logic} & \multicolumn{1}{c}{MMMU-Pro} & \multicolumn{1}{c}{MathVerse} & \multicolumn{1}{c}{Math$^{\text{Text}}$} & \multicolumn{1}{c|}{Coding$^{\text{Text}}$} & \multicolumn{1}{c}{Average}\\
\midrule
Gemini-2.0 & 21.3 & 50.0 & 70.0 & 50.0 & 50.0 & 32.5 \\
Claude-3.5 & \textbf{37.5} & 60.0 & 80.0 & 70.0 & 70.0 & \textbf{48.3}\\
OpenAI-o1 & 28.8 & 60.0 & \textbf{90.0} & \textbf{90.0} & \textbf{90.0} & 46.7\\
GPT-4o & 25.0 & \textbf{70.0} & 80.0 & 60.0 & 50.0 & 38.3\\
\bottomrule
\end{tabular}
}
\caption{\textbf{Human Evaluation Results across LMMs on \dataset{}.} Math$^{\text{Text}}$ and Coding$^{\text{Text}}$ represent two text-only task categories. The scores represent the average percentage of correct samples among all samples.}
\label{tab:humanacc}
\end{table*}

\begin{table*}[!t]
\centering
\renewcommand{\arraystretch}{1}
\scalebox{0.8}{
\begin{tabular}{l|c|ccccc|c}
\toprule
\multirow{1}{*}{Model} & \multirow{1}{*}{\# Round} & \multicolumn{1}{c}{Visual Logic} & \multicolumn{1}{c}{MMMU-Pro} & \multicolumn{1}{c}{MathVerse} & \multicolumn{1}{c}{Math$^{\text{Text}}$} & \multicolumn{1}{c|}{Coding$^{\text{Text}}$} & \multicolumn{1}{c}{Average}\\
\midrule
\multirow{3}{*}{Gemini-2.0} & 1 & 38.1 & 20.0 & 33.3 & 0.0 & 80.0 & 37.0 \\
& 2 & 20.6 & 0.0 & 33.3 & 20.0 & 20.0 & 19.8 \\
& 3 & 41.3 & 80.0 & 33.3 & 80.0 & 0.0 & 43.2\\
\midrule
\multirow{3}{*}{Claude-3.5} & 1 & 38.0 & 0.0 & 50.0 & 33.3 & 66.7 & 37.1\\
& 2 & 32.0 & 25.0 & 50.0 & 33.3 & 66.7 & 30.6\\
& 3 & 30.0 & 75.0 & 0.0 & 66.7 & 0.0 & 32.3\\
\midrule
\multirow{3}{*}{OpenAI-o1} & 1 & 38.6 & 0.0 & 100.0 & 11.1 & 100.0 & 39.1\\
& 2 & 21.1 & 0.0 & 0.0 & 0.0 & 0.0 & 18.8\\
& 3 & 40.4 & 100.0 & 0.0 & 0.0 & 0.0 & 42.2\\
\midrule
\multirow{3}{*}{GPT-4o} & 1 & 41.7 & 33.3 & 100.0 & 25.0 & 40.0 & 41.9\\
& 2 & 31.7 & 0.0 & 0.0 & 0.0 & 0.0 & 25.7\\
& 3 & 26.7 & 66.7 & 0.0 & 75.0 & 60.0 & 32.4\\
\bottomrule
\end{tabular}
}
\caption{\textbf{Correction Rate Results across various LMMs on \dataset{}.} Math$^{\text{Text}}$ and Coding$^{\text{Text}}$ represent two text-only task categories. \# Round denotes the number of interaction rounds. The correction rate is the percentage of corrected samples among all erroneous samples.}
\label{tab:humancorrectrate}
\end{table*}

\begin{figure*}[t]
    \centering
    \begin{minipage}[t]{0.22\linewidth}
        \resizebox{\linewidth}{!}{\includegraphics{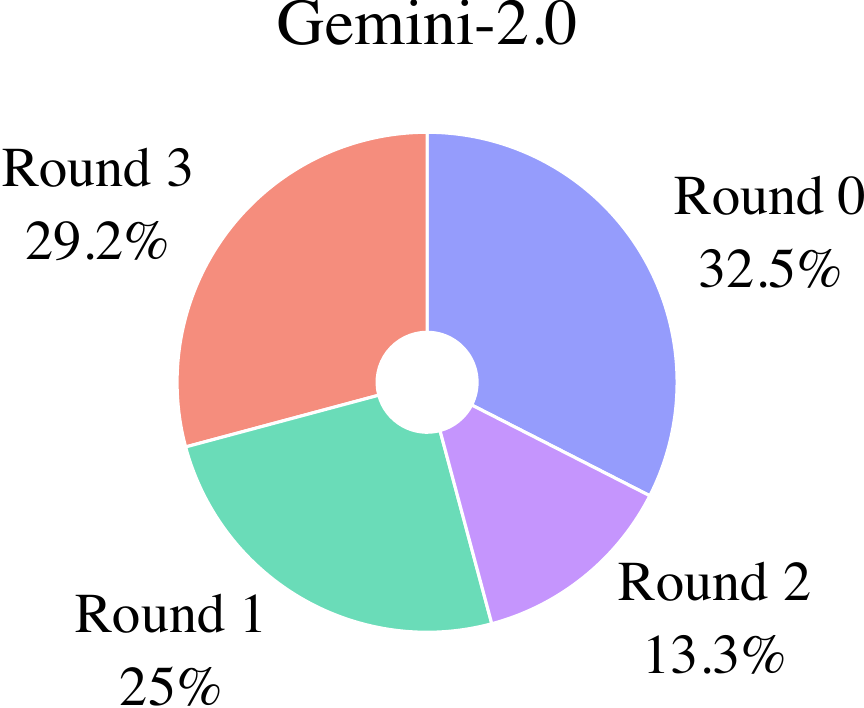}}
    \end{minipage}
    \hfill
    \begin{minipage}[t]{0.22\linewidth}
        \resizebox{\linewidth}{!}{\includegraphics{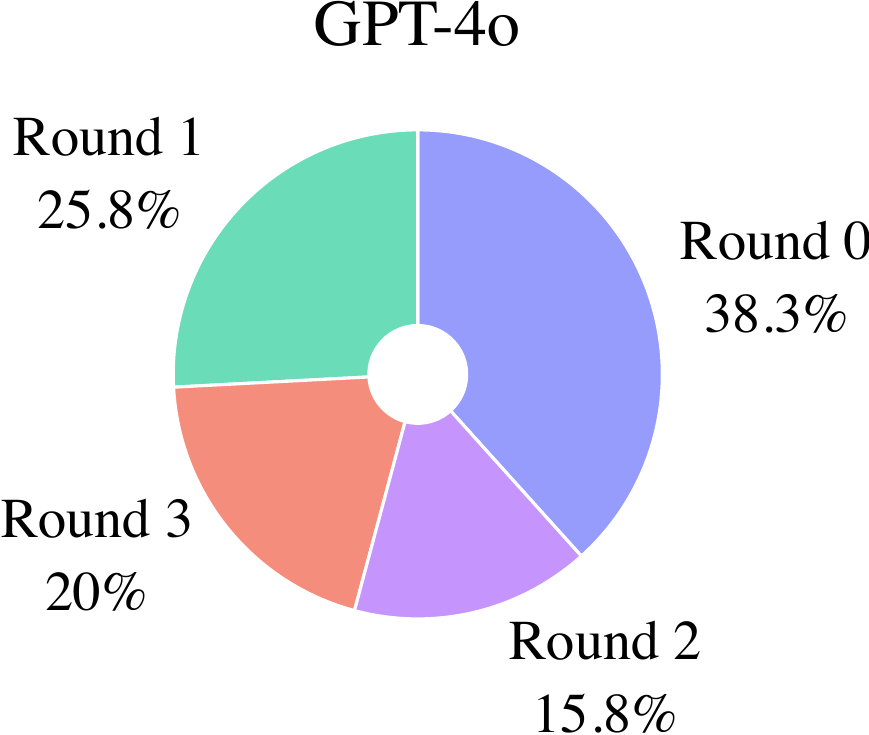}}
    \end{minipage}
    \hfill
    \begin{minipage}[t]{0.22\linewidth}
        \resizebox{\linewidth}{!}{\includegraphics{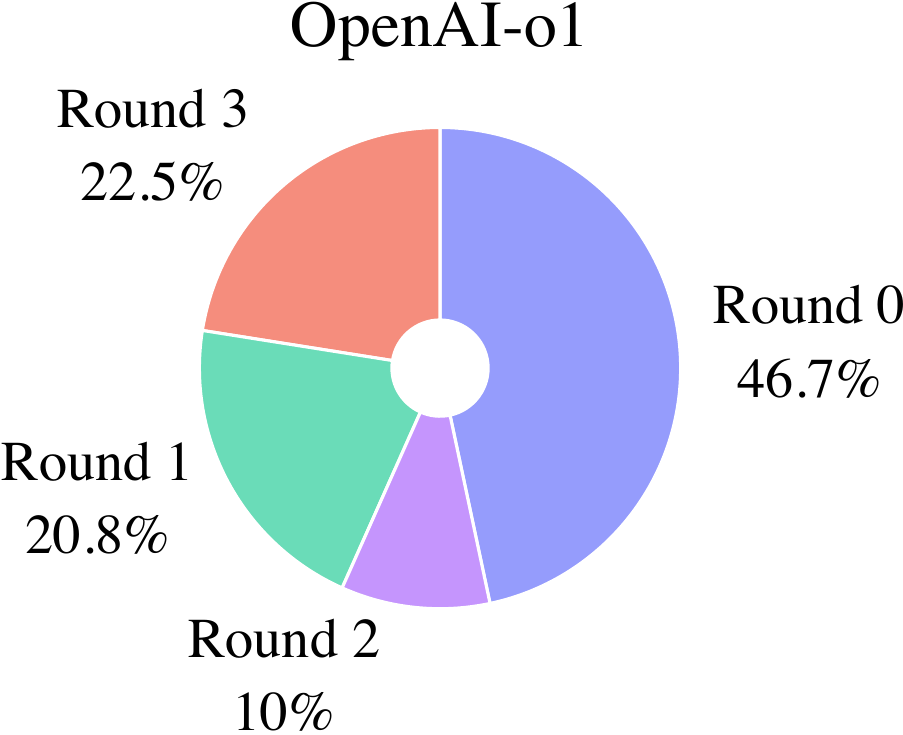}}
    \end{minipage}
    \hfill
    \begin{minipage}[t]{0.22\linewidth}
        \resizebox{\linewidth}{!}{\includegraphics{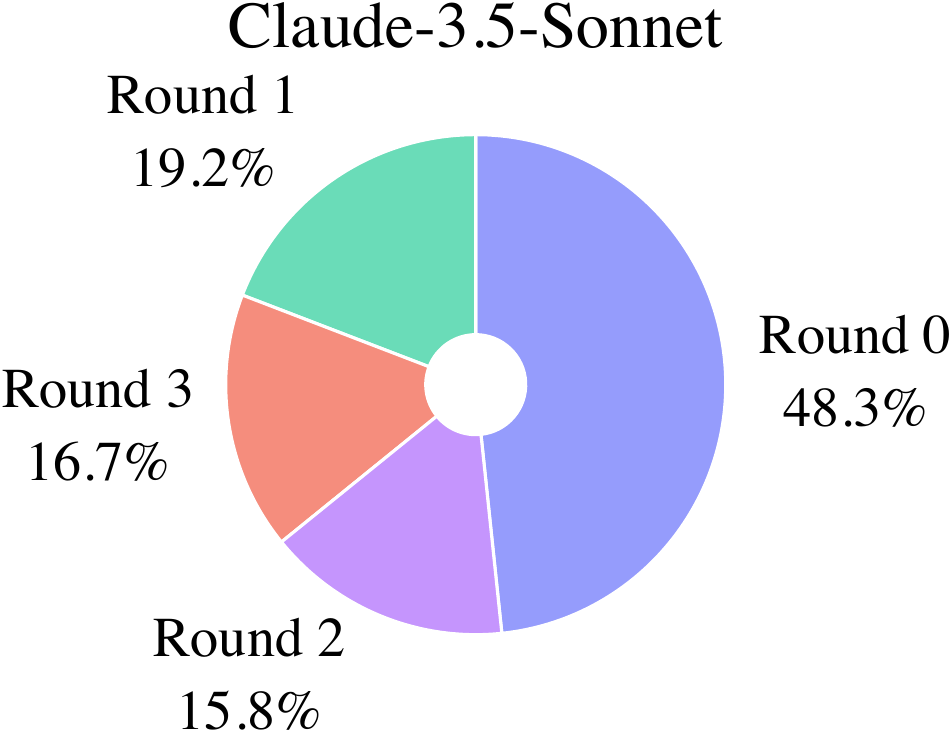}}
    \end{minipage}
    \caption{Distribution of samples being corrected in each round. We can observe that Claude-3.5-Sonnet archives the best performance in round 0.}
    \label{fig:piedistribution}
\end{figure*}

\begin{figure}[t]
    \centering
    \includegraphics[width=0.98\linewidth]{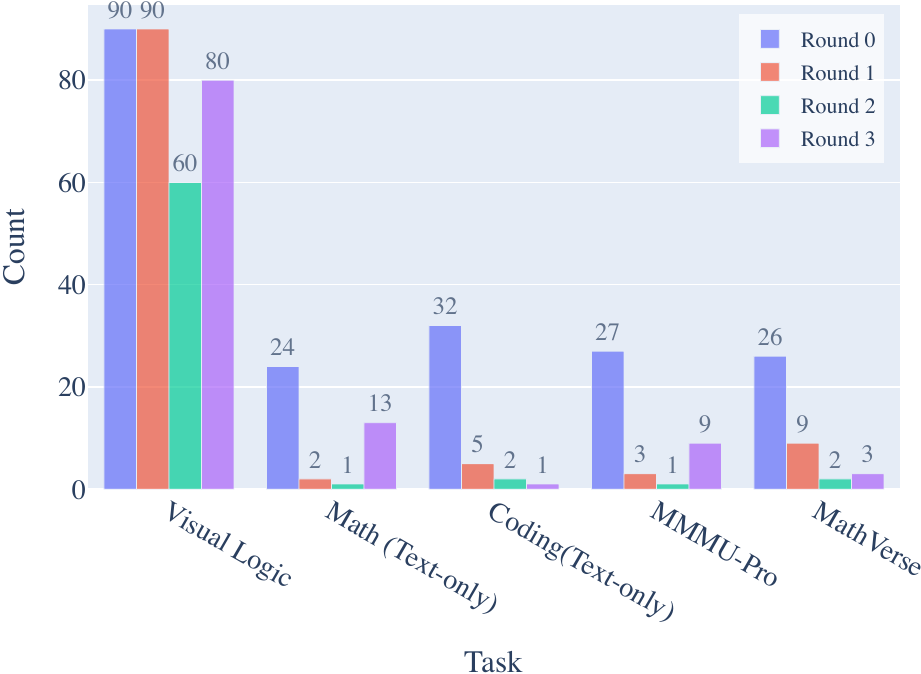}
    \caption{Distribution of corrected samples across various task categories. Visual logic tasks are mostly resolved within the first two rounds, whereas Math (Text-only) and MMMU-Pro tasks show few corrections in rounds 1 and 2. In contrast, Coding (Text-only) and MathVerse tasks exhibit corrections during rounds 1 and 2.}
    \label{fig:taskdist}
\end{figure}

\textbf{Current LMMs struggle to enhance performance through feedback.} As shown in the tables, most LMMs are unable to correct all erroneous samples, even when provided with feedback from state-of-the-art proprietary models such as Claude-Sonnet-4 and GPT-4o. For example, consider the two SOTA open-source models, Qwen2.5-VL-7B and Molmo. Qwen2.5-VL-7B achieves a 69.2\% correction rate on the MathVerse dataset with GPT-4o's feedback and 39.6\% correction rate on the MMMU-Pro dataset. Similarly, Molmo-7B attains correction rates of 55.1\% and 51.7\% on the MathVerse and MMMU-Pro datasets, respectively. Overall, the correction rates for the rest of the models remain below 50\%, no matter which feedback providers. This suggests that even with constructive feedback from proprietary LMMs, current models struggle to enhance performance through feedback generally.

\textbf{Accuracy result may not truly reflect the model’s capability to improve itself from feedback.} As shown in Table~\ref{tab:mathverse}, although InternVL2-8B achieves a higher accuracy (38.1\%), its correction rate is only 49.6\%. In contrast, Qwen2-VL-7B, with a lower accuracy of 22.5\%, attains the highest correction rate of 66.8\% when using GPT-4o's feedback. Similarly, Molmo-7B surpasses InternVL2-8B in correction rate despite having lower accuracy. On the MMMU-Pro dataset (see Table~\ref{tab:mmmupro}), LLaVA-OneVision-7B records 47.1\% but only a 31.7\% correction rate, which is lower than that of several models that have inferior accuracy (e.g., InternVL2-8B, Molmo-7B, GLM-4v-9B, and Phi3.5-Vision-4.2B). This inconsistency between initial answering ability and self-improvement capability indicates that evaluating models solely on accuracy may not fully capture their true potential.

\textbf{Simple feedback also enhances performance.} In addition to using detailed LMM-generated feedback, we evaluated models with binary (0/1) feedback that simply indicates the correctness of their current response. Surprisingly, the results show that all models benefit from this simple feedback mechanism. This suggests that while LMMs have the inherent potential to generate correct answers, they may require additional prompting techniques to fully harness their problem-solving capabilities.

\textbf{LMM-generated feedback is not always better than simple feedback.} By comparing the results obtained using \emph{Detail} feedback from GPT-4o with those using \emph{Simple} binary feedback, we observe that most models perform better with detailed feedback. For example, on the MathVerse dataset, LLaVA-OneVision-7B achieves 36.2\% with detailed feedback versus 18.0\% with binary feedback; InternVL2-8B increases from 41.2\% to 49.6\%; and MiniCPM-V increases from 20.3\% to 28.4\%. However, Qwen2-VL scores 66.8\% with detailed feedback and 72.2\% with simple feedback. Similarly, on the MMMU-Pro dataset, Fuyu-8B performs worse with detailed feedback (6.0\% vs. 8.7\%).

\textbf{The quality of feedback is crucial: low-quality feedback can degrade performance more than simply providing binary (0/1) feedback.} We compare the feedback provided by GPT-4o and Gemini-1.5-Flash on the challenging MathVerse dataset, where most models achieve accuracies below 30\%, highlighting the difficulty of its problem instances. We find that delivering simple binary feedback that merely indicates the correctness of the tested model's output can outperform LMM-generated detailed feedback (Gemini-1.5-Flash). Specifically, the correction rates using simple feedback exceed those with detailed feedback for several models: Molmo-7B (38.9\% vs. 36.5\%), MiniCPM-V (25.6\% vs. 16.6\%), Phi3.5-Vision-4.2B (33.7\% vs. 31.3\%), and Qwen2-VL-7B (44.9\% vs. 41.9\%).

\subsection{Experimental Analysis on Human Benchmarking}

In this section, we will introduce the human evaluation results of several well-known closed-source families: OpenAI (\texttt{GPT-4o}, \texttt{OpenAI-o1}), Claude (\texttt{Claude-3.5-Sonnet-20241022}), and Gemini (\texttt{Gemini-2.0-Flash-Exp}).

\textbf{Overall Accuracy Results.} In Table~\ref{tab:humanacc}: (1) The best scores for each subcategory in our \dataset{} are 37.5\% (Claude-3.5-Sonnet), 70.0\% (GPT-4o), 90\% (OpenAI-o1), and 90\% (OpenAI-o1), respectively. (2) Overall, Claude-3.5 achieves the highest average accuracy at 48.3\%.

\textbf{Correction rate results analysis.} Comparing the correction rates across rounds in Table \ref{tab:humancorrectrate} reveals that GPT-4o benefits the most from human feedback in the first round, correcting 41.9\% of erroneous samples, while Claude-3.5 exhibits its strongest correction performance in the second round, with 30.6\% of erroneous samples corrected. Given that the ground truth answer is provided in the third round, all LMMs are able to supply their reasoning steps for selecting the correct answer.

\textbf{Distribution of Tasks Corrected Across Rounds.} Figure~\ref{fig:piedistribution} illustrates the distribution of tasks solved by each LMM across the interaction rounds. Round 0 represents the initial accuracy before beginning human-AI interactions. For example, GPT-4o solved 38.3\% of instances in Round 0, 25.8\% in Round 1, and 20\% in Round 2. Additionally, during the first two rounds, both OpenAI-o1 and Claude-3.5-Sonnet solved the same number of samples, achieving a performance of 67.5\%.

\textbf{Distribution of corrected samples across various task categories.} As shown in Figure \ref{fig:taskdist}, Visual logic tasks are mostly resolved within the first two rounds, whereas Math (Text-only) and MMMU-Pro tasks show few corrections in rounds 1 and 2. In contrast, Coding (Text-only) and MathVerse tasks exhibit corrections during rounds 1 and 2.

\section*{Conclusion}
In this work, we introduced \bench, the first solution to concern the critical importance of evaluating the interactive intelligence of current LMMs. We build an interactive framework \agent{} which can be applied to any LMM and dataset to bootstrap the testing in an interactive way. We conduct the comprehensive evaluations on 10 open-source LMMs by demonstrating with two representative datasets MathVerse and MMMU-Pro. Additionally, we present \dataset{}, a new benchmark for manually testing the proprietary models such as OpenAI-o1 and Claude-3.5 with 120 curated samples. Our evaluation results show that even the SOTA LMM (like OpenAI-o1) can only correct their results through human feedback with less than 50\%. Several findings point to the essential need for methods that improve the LMM's ability to receive feedback to improve itself.

\section*{Limitations}
Our method is not without limitations.
First, as an initial attempt in the field, this work proposes a straightforward method to bootstrap the LMMs in an interactive way. We use the proprietary LMM to simulate humans, mimicking the human-AI interaction process. Due to the difficulty of existing benchmarks, the proprietary LMMs may not fully provide all pertinent feedback, though we propose two strategies: 1) select the intersection set for testing, and 2) record the valid output only. Due to the limitation of GPU memory, we have to select the tested LMMs within 7B parameters.

\section*{Acknowledgement}

This project is supported by Mike Zheng Shou's Start-Up Grant (A-0009453-03-00).

\bibliography{main}

\newpage
\appendix

\begin{table*}[!h]
\centering
\renewcommand{\arraystretch}{1.3}
\scalebox{0.8}{
\begin{tabular}{l|l|l}
\toprule
Model & Release Time & \multicolumn{1}{c}{Source} \\
\midrule
\multicolumn{3}{c}{\textit{Proprietary Models}} \\
\cmidrule{1-3}
GPT-4o \cite{gpt4o} & 2024-08-26 & \url{https://openai.com/index/hello-gpt-4o/}\\
OpenAI-o1 \cite{openaio1} & 2024-12-17 & \url{https://openai.com/o1/} \\
Gemini-1.5-Flash \cite{gemini-1.5} & 2024-09-24 & \url{https://deepmind.google/technologies/gemini/} \\
Gemini-2.0-Flash & 2025-01-21 & \url{https://deepmind.google/technologies/gemini/}\\
Claude-3.5-Sonnet & 2024-10-22 & \url{https://www.anthropic.com/claude/sonnet} \\
Claude-Sonnet-4 & 2025-05-23 & \url{https://www.anthropic.com/news/claude-4} \\
\midrule
\multicolumn{3}{c}{\textit{Open-source Models}} \\
\cmidrule{1-3}
LLaVA-One-Vision & 2024-08-05 & \url{https://llava-vl.github.io/blog/2024-08-05-llava-onevision/} \\
InterVL2-8B & 2024-07-04 & \url{https://internvl.github.io/blog/2024-07-02-InternVL-2.0}\\
InterVL3-8B & 2025-04-11 & \url{https://huggingface.co/OpenGVLab/InternVL3-8B}\\
Molmo-7B & 2024-09-24 & \url{https://huggingface.co/allenai/Molmo-7B-D-0924} \\
MiniCPM-V & 2024-08-03 & \url{https://huggingface.co/openbmb/MiniCPM-V} \\
GLM-4V-9B & 2024-11-01 & \url{https://huggingface.co/THUDM/glm-4v-9b}\\
Pih3.5-Vision-4.2B & 2024-08-20 & \url{https://huggingface.co/microsoft/Phi-3.5-vision-instruct}\\
LLaVA-1.5-7B & 2023-10-05 & \url{https://huggingface.co/liuhaotian/llava-v1.5-7b}\\
LLaVA-1.6-Mistral-7B & 2024-01-30 & \url{https://huggingface.co/llava-hf/llava-v1.6-mistral-7b-hf} \\
Fuyu-8B & 2023-10-27 & \url{https://huggingface.co/adept/fuyu-8b}\\
Qwen2-VL-7B & 2024-08-30 & \url{https://huggingface.co/Qwen/Qwen2.5-VL-7B-Instruct}\\
Qwen2.5-VL-7B & 2025-02-20 & \url{https://huggingface.co/Qwen/Qwen2.5-VL-7B-Instruct}\\
Seed-1.5-VL-Thinking & 2025-05-13 & \url{https://github.com/ByteDance-Seed/Seed1.5-VL} \\
\bottomrule
\end{tabular}
}
\caption{The release time and model source of LMMs used in our \bench{}.}
\label{tab:modelsource}
\end{table*}

\section{Model Sources.} For different LMMs, we select their latest models with sizes around 7B for evaluation. Table \ref{tab:modelsource} presents the release time and model sources of LMMs used in \bench{}.

\section{Error Analysis.}
The Table \ref{tab:enhancement-corrections} shows the answers generated after receiving feedback from two feedback providers, along with the corrected flag compared to the GT answer. Across these 10 questions, 11 out of 20 samples could not be corrected. Half of the questions remained uncorrected regardless of the feedback provider. Except for the first question in the table, the LMM generates identical answers after receiving feedback from different providers. This pattern implies that the model’s capacity to improve based on feedback resembles an inherent ability. Prompting LMM with feedback can be considered another prompting strategy to invoke the inherent ability rather than a robust ability to reason and incorporate new information to address challenging questions.

\begin{table*}[t]
\centering
\scalebox{0.6}{
\begin{tabular}{lccccc}
\hline
\textbf{Task ID} & \textbf{Enhanced by Gemini-1.5-Flash} & \textbf{Has it been corrected?} & \textbf{Enhanced by Claude-3.5-Sonnet} & \textbf{GT Answer} & \textbf{Has it been corrected?} \\
\hline
History\_134                   & B & Yes & A & B & No  \\
Pharmacy\_287                  & A & No  & A & B & No  \\
validation\_Sociology\_29      & C & No  & C & B & No  \\
Literature\_19                 & D & No  & D & B & No  \\
Agriculture\_169               & C & Yes & C & C & Yes \\
validation\_Economics\_12      & B & No  & B & A & No  \\
Electronics\_68                & A & Yes & A & A & Yes \\
Design\_126                    & D & Yes & D & D & Yes \\
Medical\_Science\_28           & A & Yes & A & A & Yes \\
test\_Pharmacy\_140            & C & No  & C & B & No  \\
\hline
\end{tabular}
}
\caption{Comparison of Enhancements and Corrections on cases from MMMU-Pro\cite{mmmupro}.}
\label{tab:enhancement-corrections}
\end{table*}

\section{Qualitative Analysis.} 
\begin{figure*}[!t]
    \centering
    \includegraphics[width=\linewidth]{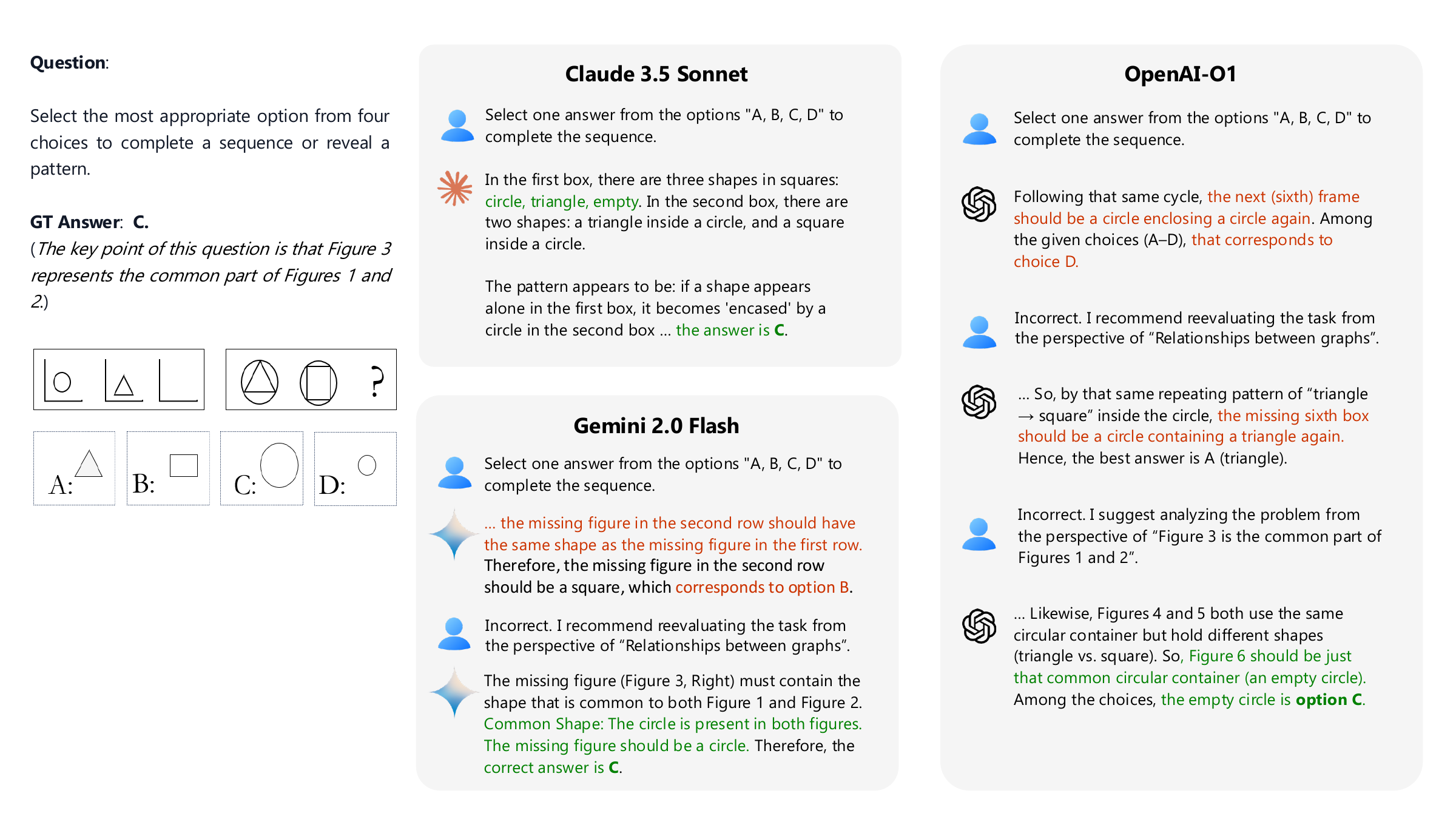}
    \caption{Qualitative results on different LMMs.}
    \label{fig:viscompare1}
\end{figure*}

\begin{figure*}[t]
    \centering
    \includegraphics[width=\linewidth]{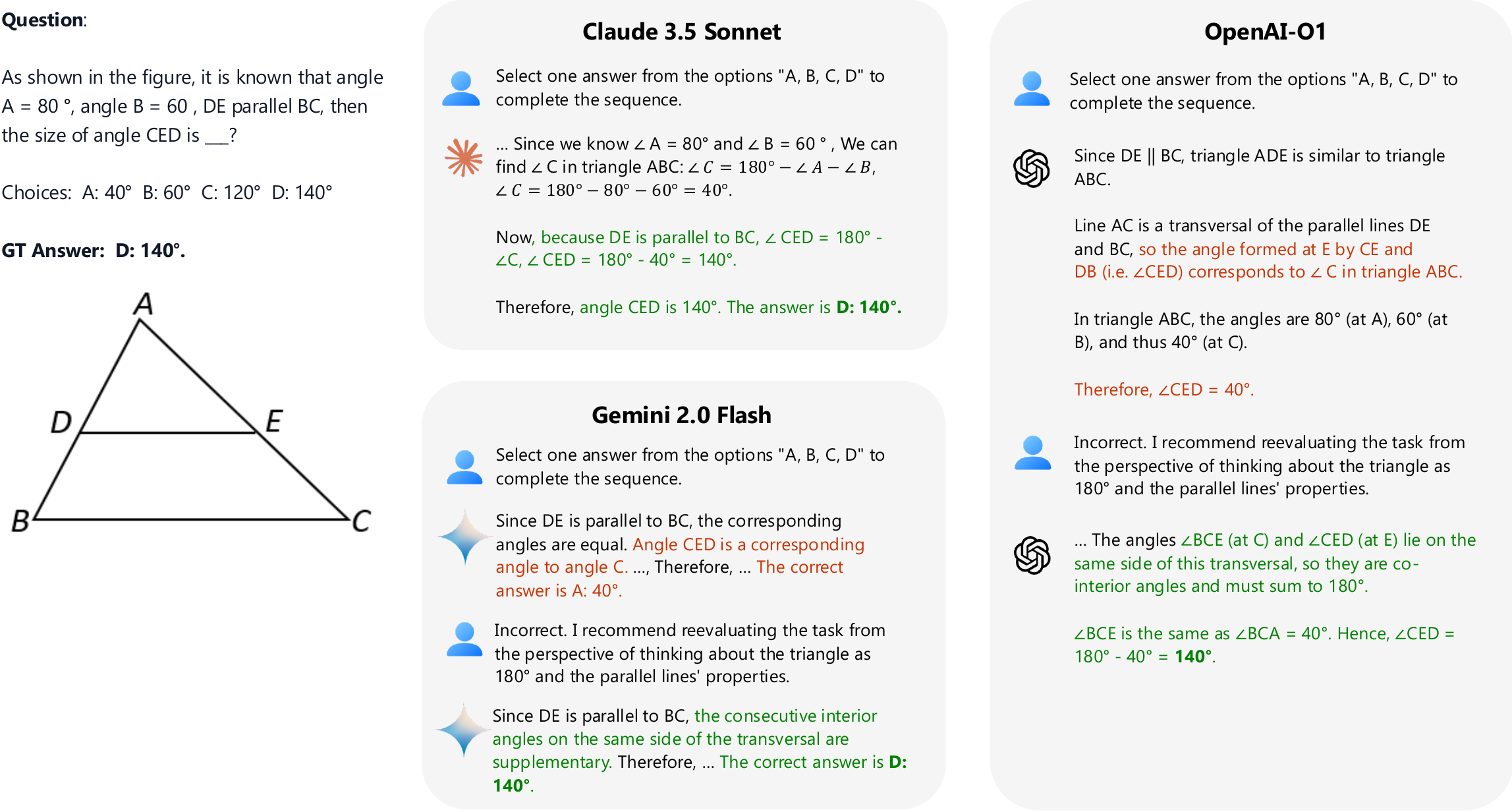}
    \caption{Qualitative results on different LMMs.}
    \label{fig:viscompare2}
\end{figure*}

\textbf{Interactive process could improve the performance of proprietary LMMs.} In Figure \ref{fig:viscompare1}, we provide the qualitative results of different models. For the same question, Claude-3.5-Sonnet gives the correct answer C without human feedback, Gemini-2.0-Flash uses two rounds while OpenAI-o1 uses three rounds. It indicates that 1) even the SOTA models like OpenAI-o1 can not fully address the visual logic problem which is worse than Claude-3.5-Sonnet, 2) the responses can be corrected by human feedback which shows that the models have the capability of interpreting and incorporating the feedback into their reasoning, 3) Different models shows a different level of this capability. Additionally, we provide another example in Figure \ref{fig:viscompare2}.

\begin{figure*}[t]
    \centering
    \includegraphics[width=\linewidth]{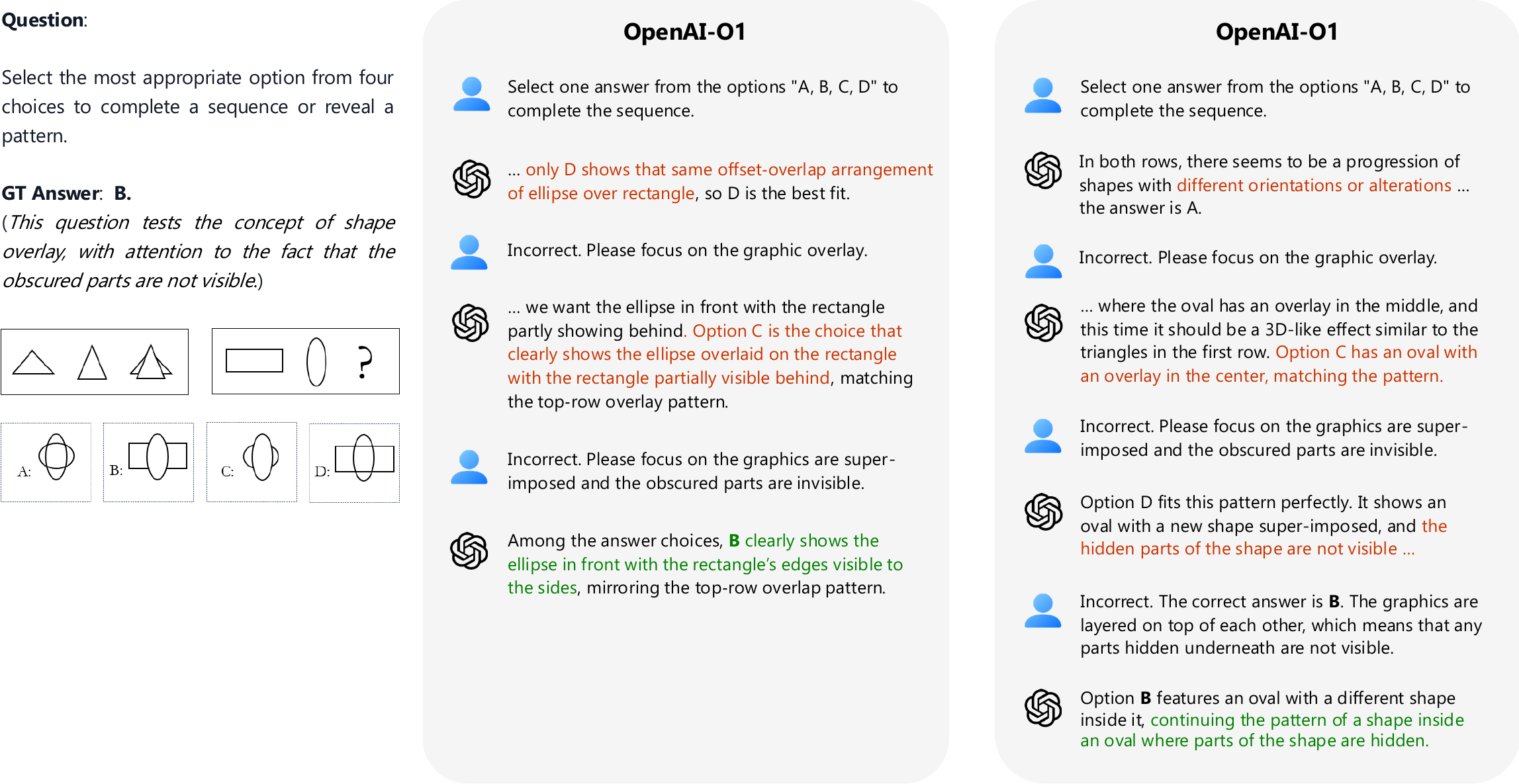}
    \caption{An example that model tends to guess answers.}
    \label{fig:visguess1}
\end{figure*}

\begin{figure*}[t]
    \centering
    \includegraphics[width=\linewidth]{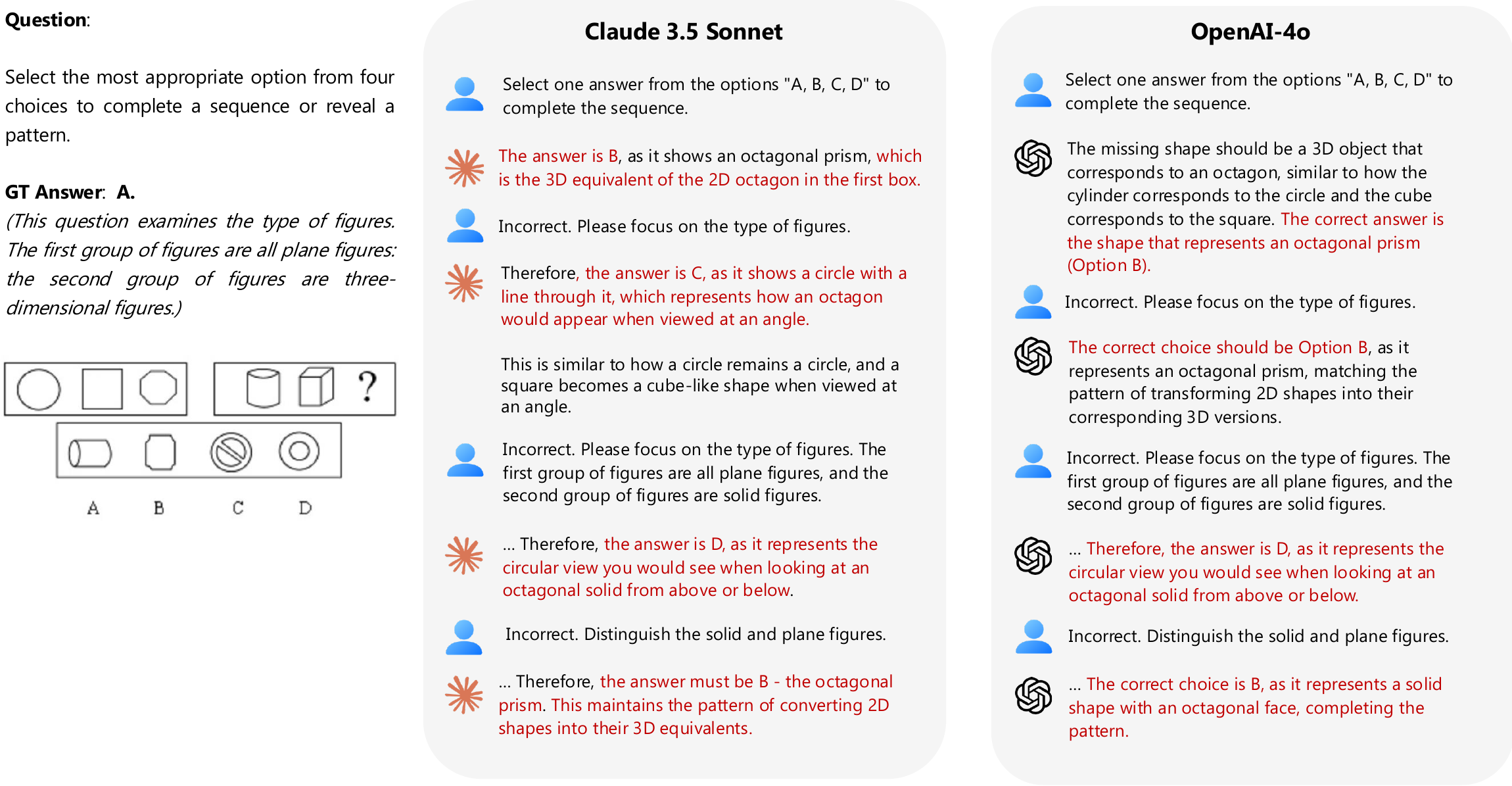}
    \caption{An example that model tends to guess answers.}
    \label{fig:visguess2}
\end{figure*}

\textbf{LMMs may not truly reasoning-They guess answers by elimination.} In Figure \ref{fig:visguess1}, we find that the model will guess the answer when we only have four options, the model tends to guess answers. For the same question, we conduct twice runs and find that OpenAI-o1 could not solve this problem at the beginning, but two different answers were given in these two runs. In the first run, the model outputs D at the beginning, while in the second run, the model outputs the A at the beginning. In the following rounds, we provide the same prompts to ensure the fairness comparison, one can see that based on the same prompt, it outputs the same answer C in the second round. The left run in the figure shows the correct answer in the third round while the right run in the figure shows the incorrect answer D. We continue to give the third feedback for round 4, and the right run finally gives answer B. It is obvious that when a problem cannot solved by a model, it will 1) outcome answer randomly, and 2) outcome the answer through an elimination approach. These results may indicate that LMMs may not always truly reason they may give the answer by guessing. Additionally, we provide another example in Figure \ref{fig:visguess2} to illustrate that LMMs may guess answers when they can not solve the challenging problems.

\begin{figure*}[t]
    \centering
    \includegraphics[width=\linewidth]{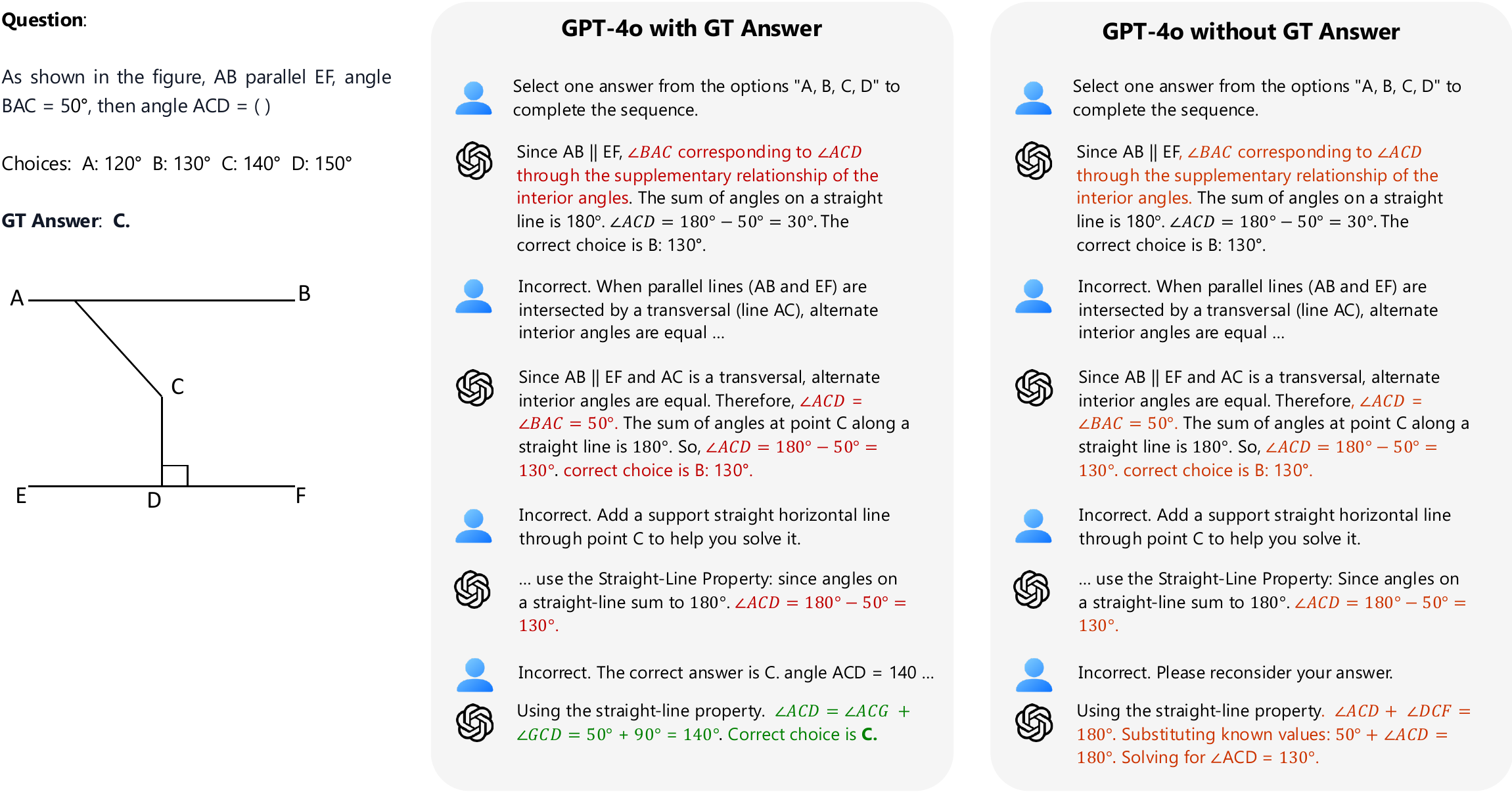}
    \caption{Qualitative results by removing GT answer in level 3 feedback.}
    \label{fig:withoutgtanswer}
\end{figure*}

\textbf{LMMs still fail when the GT answer is not provided in the level 3 feedback.} As discussed in Section \ref{sec:humanbench}, we include the GT answer in the level 3 feedback prompt to examine whether the model can generate the correct reasoning procedure that leads to the correct answer. When we remove the GT answer as in Figure \ref{fig:withoutgtanswer}, the model still fails to produce the correct answer, indicating its limited capability in solving challenging problems even when detailed feedback is provided as guidance.

\section{Examples of Feedback.} We provide the examples of feedback provided by Claude-3.5-Sonnet on MathVerse and MMMU-Pro, respectively. As these examples show that after providing the feedback, the questions are solved correctly, and the provided feedback is concise and pertinent without leaking the GT. The feedback are mainly focused on analyzing the question and providing the reasoning thoughts for the tested LMMs to use the additional information for solving questions.

\begin{tcolorbox}[colback=black!4!white,colframe=black!70!white]
\textbf{\textbf{prompt\_system}}= """You are a reasoning assistant tasked with solving questions based on visual analysis. Your goal is to analyze the provided question and image carefully and provide the correct answer.

Instructions:
1. Focus on observing all relevant details in the image.
2. Ensure thorough and logical reasoning before arriving at a conclusion.
3. Respond concisely by selecting only a single letter: A, B, C, or D.

Remember, no explanations or additional text are needed—only the chosen letter as the answer."""
\end{tcolorbox}

\begin{tcolorbox}[colback=black!4!white,colframe=black!70!white]
\textbf{prompt\_template} = """
Please carefully analyze the image and choose the correct answer to the following question.

\{question\}

Take your time to observe all relevant details in the image and ensure your reasoning is thorough before selecting your answer.

Answer with only a single letter (A, B, C, or D):
"""
\end{tcolorbox}

\begin{tcolorbox}[colback=black!4!white,colframe=black!70!white]
\textbf{prompt\_template\_local} = """
You are a reasoning assistant tasked with solving questions based on visual analysis. Your goal is to analyze the question and image carefully and provide the correct answer.

Instructions:

1. Focus on observing all relevant details in the image.

2. Ensure thorough and logical reasoning before arriving at a conclusion.

\{question\}

Answer with only a single letter (A, B, C, or D):
"""
\end{tcolorbox}

\begin{tcolorbox}[colback=black!4!white,colframe=black!70!white]
\textbf{prompt\_feedback\_system} = You are tasked with generating constructive feedback when solving questions.

Instructions:

1. Analyze the provided question and the incorrect answer.

2. Identify general reasoning flaws or oversights that might have led to the incorrect response.

3. Suggest how the reasoning can be improved, such as paying more attention to specific observations, question details, or logical deductions.

4. DO NOT mention the correct answer or any specific details of the question.

Your feedback should be constructive, and helpful.

\end{tcolorbox}

\begin{tcolorbox}[colback=black!4!white,colframe=black!70!white]
\textbf{prompt\_feedback} = Question: \{question\}

Previous answer: '\{predicted\_answer\}' was incorrect.

Provide a feedback that helps correctly answer the question.
\end{tcolorbox}

\begin{tcolorbox}[colback=black!4!white,colframe=black!70!white]
\textbf{prompt\_with\_feedback} = """
You are an assistant tasked with re-evaluating a question after receiving feedback on a previous incorrect answer.

Instructions:
1. Carefully review the question and the provided feedback.

2. Use the feedback to refine your reasoning and identify details or logical steps you may have overlooked.

3. Re-examine all aspects of the question thoroughly before selecting your answer.

{question}

Your previously selected answer: \{predicted\_answer\} was incorrect.

Feedback: \{feedback\}

Answer with only a single letter (A, B, C, or D):
"""
\end{tcolorbox}

\begin{tcolorbox}[colback=black!4!white,colframe=black!70!white]
\textbf{prompt\_with\_simple\_feedback} = """
You are an assistant tasked with re-evaluating a question on a previous incorrect answer.

Instructions:

1. Carefully review the question.

2. Refine your reasoning and identify details or logical steps you may have overlooked.

3. Re-examine all aspects of the question thoroughly before selecting your answer.

\{question\}

Your previously selected answer: \{predicted\_answer\} was incorrect.

Answer with only a single letter (A, B, C, or D):
"""
\end{tcolorbox}

\begin{table*}[]
    \centering
    \scalebox{0.8}{
    \begin{tabular}{lccccp{6cm}}
    \toprule
    ID & Initial Answer & New Answer(simple) & New Answer(Detail) & GT & Feedback \\
    \midrule
        73 & A & B & D & D & "Let me provide some constructive feedback to help with this geometry problem:

1. When analyzing angles formed by intersecting lines, remember these key principles:
- Adjacent angles sum to 180°
- When one line rotates, it affects multiple angles simultaneously
- The rotation angle is related to the difference between initial and final positions

2. Common mistakes to avoid:
- Don't assume the rotation angle is the same as the given angle
- Consider what the final position needs to be (parallel to line a)
- Think about the smallest rotation needed to achieve the desired result

3. Suggested approach:
- First identify the target position (where line b needs to end up)
- Look at the relationship between angle 1 and the amount of rotation needed
- Consider that the rotation should make the lines parallel
- Calculate how much movement is needed from the current position to the final position \\
\hline
102& A&B&D&D&"Here's some constructive feedback to help solve this problem:

1. When analyzing angles formed by intersecting lines, remember to look for:
   - Corresponding angles
   - Alternate interior angles
   - Alternate exterior angles
   - Supplementary angles (angles that add up to 180°)

2. In this case, pay special attention to:
   - The relationship between angle 4 and the other angles
   - The fact that angle 1 and angle 2 are on the same side of a line
   - How these angles relate to form a complete picture

3. Key points to improve your approach:
   - Draw out all the angle relationships clearly
   - Remember that when two parallel lines are cut by a transversal, certain angle relationships are formed
   - Use the given information that angles 1, 2, and 3 are all equal to 55°"\\
    
    \bottomrule
    \end{tabular}}
    \caption{Feedback examples provided by Claude-3.5-Sonnet on MathVerse Dataset.}
    \label{tab:feedbackmath}
\end{table*}

\begin{table*}[]
    \centering
    \scalebox{0.72}{
    \begin{tabular}{lccccp{6cm}}
    \toprule
    ID & Initial Answer & New Answer(simple) & New Answer(Detail) & GT & Feedback \\
    \midrule
        test\_Pharmacy\_287&A&B&B&B&"Here's some constructive feedback to help with answering this type of question:

1. When analyzing protein structure questions, carefully examine the image for key structural characteristics:
- Look for distinct individual polypeptide chains/subunits
- Notice how the helices are arranged and connected
- Consider whether the structure appears to be a single unit or multiple units working together

2. Remember the definition of quaternary structure:
- Review what specifically defines quaternary structure in proteins
- Don't confuse complex folding patterns within a single chain with quaternary structure

3. Methodology for improvement:
- First identify all visible protein chains in the diagram
- Look for points of interaction between separate chains
- Don't assume that just because a protein structure looks complex, it automatically has quaternary structure
 \\
\hline
test\_Math\_270&C&A&A&A&"Here's some constructive feedback to help solve this problem:

1. Approach Organization:
- Before diving into calculations, take a moment to visualize what's happening: a ladder sliding down a wall with its bottom moving outward at a constant rate
- Consider what geometric principles are involved (right triangle, Pythagorean theorem)
- Think about what information is specifically asked for (rate of change of height)

2. Key Considerations:
- Pay attention to the specific moment you're asked about (when the top reaches the wall height)
- Remember that this is a related rates problem where one rate is given (bottom moving out) and another is asked for (top moving down)
- Consider drawing the triangle at the specific instant mentioned in the question

3. Problem-Solving Strategy:
- Set up your equation carefully before plugging in numbers
- When dealing with rates of change, differentiation will be needed\\
    
    \bottomrule
    \end{tabular}}
    \caption{Feedback examples provided by Claude-3.5-Sonnet on MMMU-Pro Dataset.}
    \label{tab:feedbackmmmupro}
\end{table*}

\end{document}